\documentclass[10pt,twocolumn,letterpaper]{article}

\usepackage{cvpr}
\usepackage{times}
\usepackage{epsfig}
\usepackage{graphicx}
\usepackage{amsmath}
\usepackage{amssymb}

\usepackage[breaklinks=true,bookmarks=false]{hyperref}

\def\vec#1{\mathchoice{\mbox{\boldmath$\displaystyle#1$}}
  {\mbox{\boldmath$\textstyle#1$}}
  {\mbox{\boldmath$\scriptstyle#1$}}
  {\mbox{\boldmath$\scriptscriptstyle#1$}}}
\newcommand{\source} {{\scriptscriptstyle \mathcal{S}}}
\newcommand{\target} {{\scriptscriptstyle \mathcal{T}}}
\newcommand{\reenact}{{\scriptscriptstyle \mathcal{R}}}

\cvprfinalcopy % *** Uncomment this line for the final submission

\def\cvprPaperID{230} % *** Enter the CVPR Paper ID here

\begin{document}

%%%%%%%%% TITLE
\title{Automatic Face Reenactment}

\author{
Pablo Garrido$^1$ \hspace{.5cm} 
Levi Valgaerts$^1$ \hspace{.5cm} 
Ole Rehmsen$^1$ \hspace{.5cm} 
Thorsten Thorm\"{a}hlen$^2$ \\[.5ex]
Patrick P\'{e}rez$^3$ \hspace{.5cm} 
Christian Theobalt$^1$ \\[.5ex]
$^1$MPI for Informatics \hspace{.5cm} 
$^2$Philipps-Universit\"{a}t Marburg \hspace{.5cm} 
$^3$Technicolor
}

\maketitle
% \thispagestyle{empty}

%%%%%%%%% ABSTRACT

\begin{abstract}
  We propose an image-based, facial reenactment system that replaces
  the face of an actor in an existing target video with the face of a
  user from a source video, while preserving the original target
  performance. Our system is fully automatic and does not require a
  database of source expressions. Instead, it is able to produce
  convincing reenactment results from a short source video captured
  with an off-the-shelf camera, such as a webcam, where the user
  performs arbitrary facial gestures. Our reenactment pipeline is
  conceived as part image retrieval and part face transfer: The image
  retrieval is based on temporal clustering of target frames and a
  novel image matching metric that combines appearance and motion to
  select candidate frames from the source video, while the face
  transfer uses a 2D warping strategy that preserves the user's
  identity. Our system excels in simplicity as it does not rely on a
  3D face model, it is robust under head motion and does not require
  the source and target performance to be similar. We show convincing
  reenactment results for videos that we recorded ourselves and for
  low-quality footage taken from the Internet.
\end{abstract}

%%%%%%%%% BODY TEXT

%===============================================================================

\section{Introduction}

%% General intro
Face replacement for images \cite{Blanz04,Bitouk08} and video
\cite{Vlasic05,Alexander09,Dale11} has been studied extensively. These
techniques substitute a face or facial performance in an existing
target image or video with a different face or performance from a
source image or video, and compose a new result that looks realistic.
As a particularly challenging case, \emph{video face reenactment}
replaces a face in a video sequence, while preserving the gestures and
facial expressions of the target actor as much as possible. Since this
process requires careful frame-by-frame analysis of the facial
performance and the generation of smooth transitions between
composites, most existing techniques demand quite some manual
interaction.

%% Method intro
In this paper, we present an entirely image-based method for video
face reenactment that is fully automatic and achieves realistic
results, even for low-quality video input, such as footage recorded
with a webcam. Given an existing \emph{target sequence of an actor}
and a self-recorded \emph{source sequence of a user} performing
arbitrary face motion, our approach produces a new \emph{reenacted}
sequence showing the facial performance of the target actor, but with
the face of the user inserted in it. We adhere to the definition of
face replacement given by Dale \etal~\cite{Dale11} and only replace
the actor's inner face region, while conserving the hair, face
outline, and skin color, as well as the background and illumination of
the target video. We solve this problem in three steps:
% face tracking, face matching, and face transfer.
First, we track the user and the actor in the source and target
sequence using a 2D deformable shape model. Then, we go over the
target sequence and look in the source sequence for frames that are
both similar in facial expression and coherent over time. Finally, we
adapt the head pose and face shape of the selected source frames to
match those of the target, and blend the results in a compositing
phase.

%% Advantages of our pipeline
Our reenactment system has several important advantages: 1) Our 2D
tracking step is robust under moderate head pose changes and allows a
freedom in camera view point. As opposed to existing methods, our
system does not require that the user and the target actor share the
same pose or face the camera frontally. 2) Our matching step is
formulated as an image retrieval task, and, as a result, source and
target performances do not have to be similar or of comparable
timing. The source sequence is not an exhaustive video database, but a
single recording that the user makes of himself going through a short
series of non-predetermined facial expressions. Even in the absence of
an exact match, our system synthesizes plausible results. 3) Our face
transfer step is simple, yet effective, and does not require a 3D face
model to map source pose and texture to the target. This saves us the
laborious task of generating and tracking a personalized face model,
something that is difficult to achieve for existing, prerecorded
footage. 4)~None of the above steps needs any manual interaction:
Given a source and target video, the reenactment is created
automatically.

%% Technical contributions
We further make the following contributions: 1) We introduce a novel
distance metric for matching faces between videos, which combines both
appearance and motion information. This allows us to retrieve similar
facial expressions, while taking into account temporal continuity. 2)
We propose an approach for segmenting the target video into temporal
clusters of similar expression, which are compared against the source
sequence. This stabilizes matching and assures a more accurate image
selection. 3) A final contribution is an image-based warping strategy
that preserves facial identity as much as possible. Based on the
estimated shape, appearance is transferred by image blending.

%% Paper overview.
The paper is organized as follows: Sec.~\ref{Sec:related_work}
and~\ref{Sec:overview} discuss related work and give a brief overview
of our system. Sec.~\ref{Sec:tracking}, \ref{Sec:matching}, and
\ref{Sec:transfer} describe the three main steps in our pipeline.
% and their contributions.
In Sec.~\ref{Sec:results}, we present results and a validation for
existing and self-recorded footage, before concluding in
Sec.~\ref{Sec:conclusion}.

%===============================================================================

\section{Related Work}
\label{Sec:related_work}

Face replacement for image and video can be roughly divided into two
categories. A first category is \mbox{\emph{facial puppetry}}
\cite{Vlasic05,Theobald09,Weise09,Kemelmacher10,Saragih11,Li12}, which
aims to transfer expressions and emotions of a user (puppeteer) to a
virtual character (puppet). Such methods are used to animate digital
avatars in games, movies and video conferences. \emph{Face swapping}
methods \cite{Bregler97,Blanz04,Bitouk08,Jones08,Alexander09,Dale11},
on the other hand, try to exchange two faces in different images or
videos such that the replaced result looks sufficiently
realistic. Swapping different faces is useful for online identity
protection, while swapping the same face (or parts of it) between
different videos is interesting for dubbing, retargeting and video
montage. At the intersection of both categories lies \emph{face
  reenactment} \cite{Dale11}, which replaces an actor's face by
swapping it with that of a user, while at the same time preserving the
actor's facial expressions and emotions. Here, the original facial
performance needs to be accurately emulated (puppetry), and the new
face with different identity needs to be inserted as naturally as
possible in the original video (swapping).

Methods for face replacement in video can be further divided based on
the underlying face representation:

A first type of methods tracks a morphable 3D model of the face that
parameterizes identity, facial expressions and other nuances. Such
systems can produce accurate 3D textured meshes and can establish a
one-to-one expression mapping between source user and target actor,
thereby simplifying and speeding up expression transfer. The
generation of such a model, however, can be time consuming and is
either done by learning a detailed 3D multilinear model from example
data spanning a large variety of identities and expressions
\cite{Vlasic05,Dale11}, or by purposely building a person-specific
blend shape model from scans of a specific actor using specialized
hardware
\cite{Eisert98,Jones08,Alexander09,Weise09,Weise11}. Moreover, the
difficulty of stably tracking a 3D model over time generally
necessitates a fair amount of manual interaction.

A second type of approaches finds similarities in head pose and facial
expression between two videos solely based on image information. These
image-based methods track the face using optical flow \cite{Li12} or a
sparse set of 2D facial features \cite{Saragih11}, and often include
an image matching step to look up similar expressions in a database
\cite{Kemelmacher10,Li12}. Many image-based face replacement systems
do not allow much head motion and assume that the actors in both
videos share a similar frontal head pose \cite{Li12}. As a result,
substantial differences in pose and appearance may produce unrealistic
composites or blending artifacts. If the task is to create a new
facial animation, additional temporal coherence constraints must be
embedded in the objective to minimize possible in-between jumps along
the sequence \cite{Kemelmacher11, Berthouzoz12}.

As far as we are aware, only the 3D morphable model technique of Dale
\etal \cite{Dale11} could be used for face reenactment thus far.
Their approach uses complex 3D mesh tracking, is not fully automatic,
requires comparable source and target head poses, and was mainly
demonstrated on sequences of similar performance.  Our method, on the
other hand, is purely image-based, and thus less complex, fully
automatic, and equipped with a face tracking and transfer step that
are robust to changes in head pose. Moreover, our image retrieval step
works on source and target sequences with notably different
performances. In this respect, our method is closely related to the
work of Efros \etal. \cite{Efros03}, Kemelmacher-Shlizerman \etal
\cite{Kemelmacher10} and Li \etal \cite{Li12}. As opposed to these
works, we do not use a dedicated source database, but only a short
sequence of the user performing arbitrary expressions and head motion.
Contrary to \cite{Kemelmacher10}, we further combine appearance and
motion similarities in our matching metric to enforce temporally
coherent image look-up. Finally, we produce a proper composite of the
user's face in the target sequence, while previous works
\cite{Kemelmacher10} and \cite{Li12} only produce an, often
stop-motion-like, assembly of source frames. Berthouzoz \etal
\cite{Berthouzoz12} use hierarchical clustering to find frames of
similar expression and head pose to produce smooth transitions between
video segments, but the obtained clusters lack temporal continuity.
Expression mapping \cite{Liu01} is another related technique, which
transfers a target expression to a neutral source face. However, this
technique does not preserve the target head motion and illumination,
and has problems inside the mouth region, where teeth are not
visible. Furthermore, our image retrieval step transfers subtle
details of the user's facial expressions, which can differ between
individual users.

%===============================================================================

\section{Overview of our Face Reenactment System}
\label{Sec:overview}

Our system takes as input two videos showing facial performances of
two different persons: a \emph{source sequence} $\mathcal{S}$ of the
user, and a \emph{target sequence} $\mathcal{T}$ of an actor. The goal
is to replace the actor's inner face region with that of the user,
while preserving the target performance, scene appearance and lighting
as faithfully as possible. The result is the \emph{reenactment
  sequence} $\mathcal{R}$. The source and target video are not assumed
to depict the same performance: We can produce reenactments for
different target videos from only a single source video, which is
assumed to show the user going through a short series of random facial
expressions while facing the camera. The target sequence can be
general footage depicting a variety of expressions and head poses.

Our approach consists of three subsequent steps
(Fig.~\ref{Fig:overview}):

\begin{figure}
  \centering
  \includegraphics[width=\linewidth]{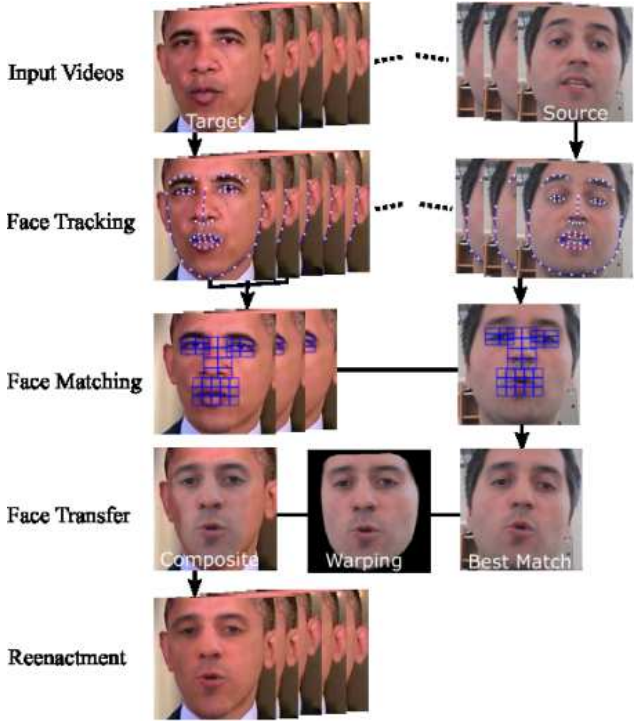}
  \caption{Overview of the proposed system.}
  \label{Fig:overview}
\end{figure}

\begin{enumerate}

\item \emph{Face Tracking} (Sec.~\ref{Sec:tracking}): A non-rigid face
  tracking algorithm tracks the user and actor throughout the videos
  and provides facial landmark points. These landmarks are stabilized
  to create a sequence of annotated frames.

\item \emph{Face Matching} (Sec.~\ref{Sec:matching}): The appearance
  of the main facial regions is encoded as a histogram of local binary
  patterns, and target and source frames are matched by a nearest
  neighbor search. This is rendered more stable by dividing the target
  sequence into chunks of similar appearance and taking into account
  face motion.

\item \emph{Face Transfer} (Sec.~\ref{Sec:transfer}): The target head
  pose is transferred to the selected source frames by warping the
  facial landmarks. A smooth transition is created by synthesizing
  in-between frames, and blending the source face into the target
  sequence using seamless cloning.

\end{enumerate}

%===============================================================================

\section{Non-Rigid Face Tracking}
\label{Sec:tracking}

To track user and actor in the source and target sequence,
respectively, we utilize a non-rigid 2D face tracking algorithm
proposed by Saragih \etal~\cite{Saragih09}, which tracks $n \! = \!
66$ consistent landmark locations on the human face (eyes, nose,
mouth, and face outline, see Fig.~\ref{Fig:region_split} (a)). The
approach is an instance of the constrained local model (CLM)
\cite{Cristinacce06}, using the subspace constrained mean-shift
algorithm as an optimization strategy. Specifically, it is based on a
3D point distribution model (PDM), which linearly models non-rigid
shape variations around 3D reference landmark locations,
$\bar{\vec{X}}_i$, $i \! = \! 1, \dots, n$, and composes them with a
global rigid transformation:
\begin{equation}
  \vec{x}_i = s P R \left( \bar{\vec{X}}_i + \Phi_i \vec{q} \right) + \vec{t} \enspace \mbox{.}
  \label{Eq:pdm}
\end{equation}
Here, $\vec{x}_i$ is the estimated 2D location of the $i$-th landmark,
and $s$, $R$, $\vec{t}$ and $\vec{q}$ the PDM parameters,
corresponding to the scaling, the 3D rotation, the 2D translation, and
the non-rigid deformation parameters. Further, $\Phi_i$ denotes the
submatrix of the basis of variation to the $i$-th landmark and $P$ is
the orthogonal projection matrix. To find the most likely landmark
locations, the algorithm uses trained local feature detectors in an
optimization framework that enforces a global prior over the combined
landmark motion. We remark that we only use the 2D landmark output
$(\vec{x}_1, ..., \vec{x}_{n})$ of the tracker, and not the underlying
3D PDM.

The facial landmarks are prone to noise and inaccuracies, especially
for expressions on which the face tracker was not trained. This can
render the face matching (see Sec.~\ref{Sec:matching}) and face
transfer (see Sec.~\ref{Sec:transfer}) less stable. To increase
tracking accuracy, we therefore employ a correction method similar to
that proposed by Garrido \etal~\cite{Garrido13}, which refines the
landmark locations using optical flow between automatically selected
key frames, i.e., frames for which the localization of the facial
features detected by the face tracker is reliable, such as a neutral
expression. To improve the smoothness of the landmark trajectories, we
do not use the estimated optical flow value at the exact landmark
location $\vec{x}_i$, like Garrido \etal, but assign a weighted
average of the flow in a circular neighborhood around
$\vec{x}_i$. This neighborhood of size $r \! \cdot \! p$ is built by
distributing $p$ points evenly on circles with radial distances of $1,
2, \dots, r$ from $\vec{x}_i$. In our experiments we choose $r \! = \!
2$ and $p \!  = \!8$, and weigh the flow values by a normalized
Gaussian centered at $\vec{x}_i$.

%===============================================================================

\section{Face Matching}
\label{Sec:matching}

A central part of our reenactment system is matching the source and
target faces under differences in head pose. Here, we find a trade-off
between exact expression matching, and temporal stability and
coherence. The tracking step of the previous section provides us with
facial landmarks which represent the face shape.  Instead of comparing
shapes directly, we match faces based on appearance and landmark
motion, depicting the facial expression and its rate of change,
respectively. Another contribution of the matching step is a temporal
clustering approach that renders the matching process more stable.

\subsection{Image Alignment and Feature Extraction}
\label{Sec:feature_extraction}

Before extracting meaningful facial features, the source and target
frames are first aligned to a common reference frame.  For this
purpose, we choose the first frame in the source sequence, which is
assumed to depict the user at rest. Unlike methods that align source
and target using a morphable 3D model \cite{Kemelmacher10}, we compute
a 2D affine transformation for each frame that optimally maps the set
of detected landmarks onto the reference shape. Since this
transformation is global, it does not change the expression in the
aligned frames. This alignment is only necessary for the temporal
clustering and frame selection of
Sec.~\ref{Sec:temporal_clustering_and_frame_selection}, but is not
applied for the subsequent steps of our system.

\begin{figure}
  \centering
  {
    \renewcommand{\tabcolsep}{1pt}
    \begin{tabular}{cc}
      \includegraphics[width=2cm]{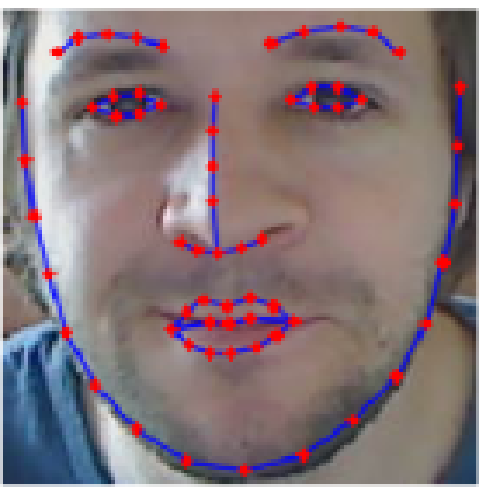} & \includegraphics[width=6cm]{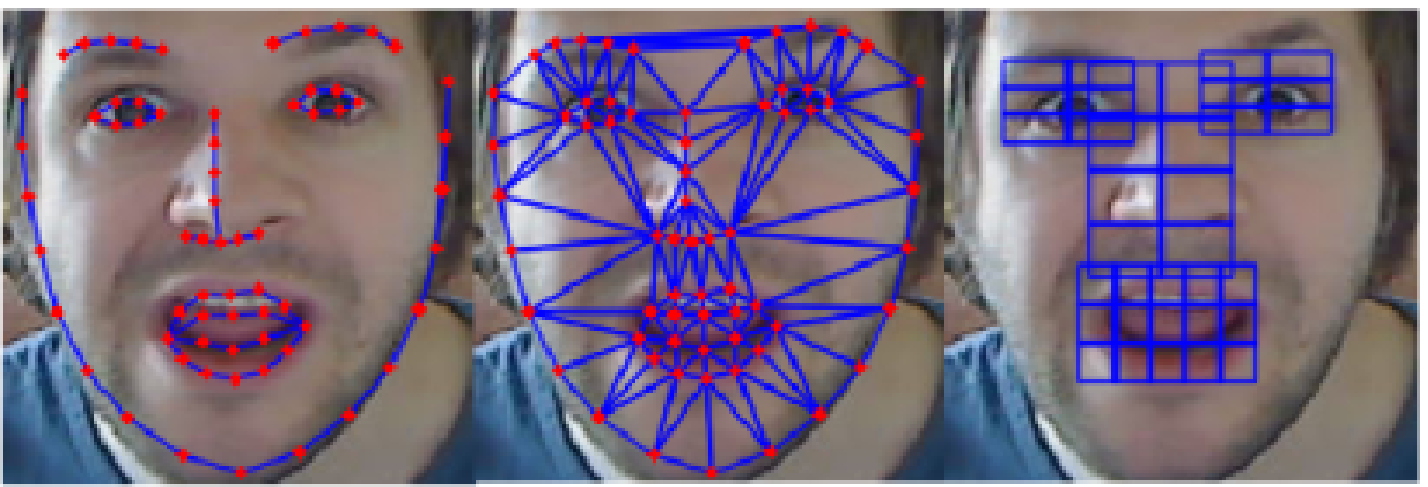} \\
      (a) & (b)
    \end{tabular}
  }
  \caption{(a) Annotated reference frame. (b) Expressive face aligned
    to the reference. Left to right: estimated landmarks,
    triangulation, and detected regions of interest. The mouth, eyes
    and nose regions are split into $3 \times 5$, $3 \times 2$ and $4
    \times 2$ tiles, respectively.}
  \label{Fig:region_split}
\end{figure}

To extract facial features in the aligned frames, we consider four
regions of interest of fixed size, which are computed as the bounding
boxes of the landmark locations corresponding to the mouth, eyes and
nose in the reference frame. After padding these regions by 10 pixels,
we split them into several tiles, as shown in
Fig.~\ref{Fig:region_split}. As feature descriptor for a region of
interest, we choose histograms of Local Binary Patterns
(LBPs)~\cite{Ojala02}, which have been found suitable for tasks such
as texture classification and face recognition
\cite{Ahonen06,Kemelmacher10,Tan10,Kemelmacher11}. An LBP encodes the
relative brightness around a given pixel by assigning a binary value
to each neighboring pixel, depending on whether its intensity is
brighter or darker. The result is an integer value for the center
pixel between 0 and $2^l$, where $l$ is the number of pixels in a
circular neighborhood. We use a uniform code \cite{Kemelmacher10},
which assigns an own label to every combination for which the number
of transitions between 0 and 1 is at most two, and a single label for
all other combinations. For a neighborhood size of $l \! = \! 8$, this
results in an LBP histogram $h$ of 59 bins for each tile. Empirically,
we found that a uniform code lacks in discriminative power to match
expressions from a wider set other than the distinctive neutral,
sadness, happiness, anger, etc. To include information at a finer
scale of detail, we additionally compute a normal LBP histogram for a
neighborhood size of $l \! = \! 4$, thereby extending $h$ to 75
bins. By concatenating the histograms for all $m$ tiles that make up a
region of interest, an LBP feature descriptor $H \! = \! \left( h_1,
  \dots, h_m \right)$ for that region is created.

\subsection{Temporal Clustering and Frame Selection}
\label{Sec:temporal_clustering_and_frame_selection}

Matching source and target frames directly may lead to abrupt
frame-to-frame expression changes in the reenactment. The reasons for
this are: 1) We experienced a sensitivity of LBP feature descriptors
w.r.t. the detected regions of interest, which can result in slightly
different source selections for similar target expressions (comparable
effects were reported by Li \etal \cite{Li12}). 2) The source sequence
is sparse and may not contain an exact match for each target
expression. 3) There is no temporal consistency in the image
selection. To overcome these shortcomings, we stabilize the matching
process by a temporal clustering approach, which finds the source
frame that is most similar to a small section of target frames.
Additionally, we enforce temporal continuity by extending the
appearance metric with a motion similarity term, which takes into
account the change in expression.

\paragraph{Temporal Clustering.}

To stabilize the selection of source frames, we divide the target
sequence into consecutive sections of similar expression and
appearance, and look for the source frame that best matches a whole
target section. To measure the similarity between two consecutive
target frames $f_\target^t, f_\target^{t+1} \in \mathcal{T}$, we
compute the \emph{appearance distance}
\begin{equation} \label{Eq:appearance_distance}
  d_{\text{app}}(f_\target^t, f_\target^{t+1}) = \sum_{j=1}^4 w_j \, d_{\chi^2} \left( H_j (f_\target^{t}), H_j (f_\target^{t+1}) \right) \enspace \mbox{,}
\end{equation}
where $H_j (f)$ is the LBP feature descriptor for the $j$-th of the
four regions of interest in $f$, $w_j$ an accompanying weight, and
$d_{\chi^2}$ the normalized chi-squared distance between two
histograms. The weights for mouth, eyes and nose regions were
experimentally set to $0.6$, $0.15$ and $0.1$, respectively.

We propose an agglomerative clustering approach that preserves
temporal continuity and proceeds hierarchically. Assuming that each
frame is initially a separate cluster, each subsequent iteration joins
the two consecutive clusters that are closest according to the metric
in Eq.  \eqref{Eq:appearance_distance}. As a linkage criterion, the
appearance distance between two consecutive clusters $\mathcal{C}_1$
and $\mathcal{C}_2$ is defined as the average of the pairwise
distances $d_{\text{app}}$ between all frames in $\mathcal{C}_1$ and
all frames in $\mathcal{C}_2$. The two clusters are merged if 1) they
only contain a single frame or 2) the variance of $d_{\text{app}}$
within the merged cluster is smaller than the maximum of the variances
within the separate clusters. The last criterion keeps the frames
within a cluster as similar as possible, and once it is not met, the
algorithm terminates. The result is a sequence of target sections
$\mathcal{C}^k$, with $k$ an index running in temporal direction over
the number of clusters. We observed that the length of a cluster
$\mathcal{C}$ varies inversely proportionally to the change in
expression and the timing of speech within $\mathcal{C}$ (see the
supplemental material for an analysis of the number of detected
clusters and their lengths).

\paragraph{Frame Selection.}

To select a source frame $f_\source^k \in \mathcal{S}$ that matches a
target section $\mathcal{C}^k$, we compute an aggregated similarity
metric over all target frames in a cluster:
\begin{equation} \label{Eq:total_distance}
  d ( \mathcal{C}, f_\source ) = 
  \sum_{f_\target \in \mathcal{C}} 
  d_{\text{app}}(f_\target, f_\source) + \tau \;
  d_{\text{mot}}(\vec{v}_{\mathcal{C}}, \vec{v}_\source )
  \enspace \mbox{.}
\end{equation}
Here, $d_{\text{app}} (f_1, f_2)$ is the appearance distance defined
in Eq.~\eqref{Eq:appearance_distance} and $d_{\text{mot}}(\vec{v}_1,
\vec{v}_2 )$ a \emph{motion distance} that measures the similarity
between two vector fields. The vector field $\vec{v}_{\mathcal{C}}$
describes the motion of the $n$ facial landmarks between two
consecutive clusters. The motion of the $i$-th landmark
$\vec{v}_{\mathcal{C} i}$ is computed as the difference of its average
positions in the current cluster $\mathcal{C}^k$ and the previous
cluster $\mathcal{C}^{k-1}$. The vector field $\vec{v}_\source$
describes the motion of the $n$ facial landmarks between two
consecutively selected source frames, i.e. for the $i$-th landmark,
$\vec{v}_{\source i}$ is the difference of its position in
$f_\source^k$ and $f_\source^{k-1}$. Note that $\vec{v}_{\mathcal{C}}$
and $\vec{v}_\source$ are computed for normalized landmark locations
in the aligned source and target frames. The motion distance
$d_{\text{mot}}$ is defined as
\begin{equation} \label{Eq:motion_distance}
  d_{\text{mot}}( \vec{v}_{\mathcal{C}}, \vec{v}_\source )
  = 
  1 - \frac{1}{3} \sum_{j = 1}^3 \exp \big(  - d_j ( \vec{v}_{\mathcal{C}}, \vec{v}_\source) \big)
  \enspace \mbox{,}
\end{equation}
where $d_1 \! \! = \! \! \frac{1}{n} \sum_i \| \vec{v}_{\mathcal{C} i}
\! - \!  \vec{v}_{\source i} \|$ measures the Euclidean distance, $d_2
\! \! = \! \! \frac{1}{n} \sum_i ( 1 - \vec{v}_{\mathcal{C} i} \cdot
\vec{v}_{\source i} / \| \vec{v}_{\mathcal{C} i} \| \|
\vec{v}_{\source i} \| )$ the angular distance, and $d_3 \!  \! = \!
\! \frac{1}{n} \sum_i | \| \vec{v}_{\mathcal{C} i} \| - \|
\vec{v}_{\source i} \| |$ the difference in magnitude between the
motion fields $\vec{v}_{\mathcal{C}}$ and $\vec{v}_\source$. The
motion distance $d_{\text{mot}}$ therefore measures how similar the
change in expression in the selected source frames is compared to the
change in expression between target clusters. It is important to
understand that consecutively selected frames $f_\source^{k-1}$ and
$f_\source^k$ do not have to be consecutive in the original source
sequence $\mathcal{S}$. Our matching metric is thus suitable for
source and target sequences that have an entirely different timing and
speed. Both the aggregated appearance distance and motion distance in
Eq.~\eqref{Eq:total_distance} are normalized to $[0,1]$ and the
weighting factor $\tau$ was set to $0.8$ for all experiments.

Given $f_\source^{k-1}$, the source frame with the minimal total
distance $d(\mathcal{C}^k,f_\source)$ over all $f_\source \in
\mathcal{S}$, is chosen as the best match $f_\source^k$ and assigned
to the central timestamp of $\mathcal{C}^k$. If $\mathcal{C}^k$
consists of a single frame, $f_\source^k$ is assigned to this
timestamp.

% ===============================================================================

\section{Face Transfer}
\label{Sec:transfer}

After selecting the best representative source frames, we transfer the
face of the user to the corresponding target frames and create the
final composite. First, we employ a 2D warping approach which
combines global and local transformations to produce a natural shape
deformation of the user's face that matches the actor in the target
sequence. The estimated shape is then utilized to transfer the user's
appearance and synthesize a compelling transition.

\subsection{Shape and Appearance Transfer}
\label{Sec:shape_and_appearance_transfer}

While only methods relying on complex 3D face models can handle large
differences in head pose between source and target \cite{Dale11}, we
present a simple, yet effective, image-based strategy that succeeds in
such cases. Inspired by work on template fitting
\cite{Blanz04,Weise09}, we formulate face transfer as a deformable 2D
shape registration that finds a user shape and pose that best
correspond to the shape and pose of the actor, while preserving the
user's identity as much as possible.

\paragraph{Shape Transfer.}

For each target frame $f_\target^t \! \in \! \mathcal{T}$, we want to
estimate the $n$ 2D landmark locations $\left( \vec{x}^t_{\reenact 1},
  ..., \vec{x}^t_{\reenact n} \right)$ of the user's face in the
reenactment sequence $\mathcal{R}$. To achieve this, we propose a
warping energy composed of two terms: a non-rigid term and an affine
term. The non-rigid term penalizes deviations from the target shape:
\begin{equation} \label{Eq:non-rigid-term}
  E_{\text{nr}} \! = \!
  \sum_{i = 1}^n \left \| \, \vec{x}^t_{\reenact i} \! - \! 
    \left( 
      \alpha_1 \, \vec{x}^{t-1}_{\target i} \! + \! 
      \alpha_2 \, \vec{x}^{t  }_{\target i} \! + \! 
      \alpha_3 \, \vec{x}^{t+1}_{\target i} 
    \right) 
  \right \|^2
  \enspace \mbox{,}
\end{equation}
where $\vec{x}^t_{\target i}$ denotes the $i$-th landmark in the
target frame at time $t$ and $\alpha_j$, $\sum_j \alpha_j \! = \! 1$,
are normalized weights (0.1, 0.8 and 0.1 in our experiments).  The
affine term penalizes deviations from the selected source shape:
\begin{equation} \label{Eq:affine-term}
  E_{\text{r}} \! = \!
  \sum_{i = 1}^n \left \| \, \vec{x}^t_{\reenact i} - 
    \left( 
      \beta_1 \, M^{k-1} \vec{x}^{k-1}_{\source i} + 
      \beta_2 \, M^{k} \vec{x}^{k}_{\source i} 
    \right) 
  \right \|^2
  \enspace \mbox{,}
\end{equation}
where $\vec{x}^{k-1}_{\source i}$ (resp.\ $\vec{x}^{k}_{\source i}$)
is the $i$-th landmark in the selected source frame immediately
preceding (resp.\ following) the current timestamp $t$, and $M$ a
global affine transformation matrix which optimally aligns the
corresponding source and target shapes. As the selected source frames
are only assigned to the central timestamp of a temporal cluster, no
selected source shape may correspond to the current target frame
$f_\target^t$, so this term effectively interpolates between the
closest selected source shapes, thereby preserving the user's
identity. The weights $\beta_j$, $\sum_j \beta_j \! = \! 1$, depend
linearly on the relative distance from $t$ to the central timestamps
of $\mathcal{C}^{k-1}$ and $\mathcal{C}^k$, being 0 or 1 if $t$
coincides with one of the cluster centers.  Combining the two terms
together with their corresponding weights $w_{\text{nr}}$ and
$w_{\text{r}}$, yields the total energy
\begin{equation} \label{Eq:total_energy}
  E_{\text{tot}}(\vec{x}^t_{\reenact i}) = w_{\text{nr}} \, E_{\text{nr}} + w_{\text{r}} \, E_{\text{r}}
  \enspace \mbox{,}
\end{equation}
where $w_{\text{nr}} + w_{\text{r}} \! = \! 1$.  A closed-form
solution to Eq.~\eqref{Eq:total_energy} for the optimal landmark
locations $\left( \vec{x}^t_{\reenact 1}, ..., \vec{x}^t_{\reenact n}
\right)$ exists.

\paragraph{Appearance Transfer.}

Once we have the optimal shape of the face in the reenactment
sequence, we transfer the appearance of the selected source frames by
inverse-warping the corresponding source texture \cite{Saragih11}
using a triangulation of the landmark points (see
Fig.~\ref{Fig:region_split} (b)). For the in-between frames, we create
a smooth transition in appearance by interpolating the texture from
the closest selected source frames using the same triangulation of the
landmarks.

\begin{figure}
  \centering
  \includegraphics[width=.9\linewidth]{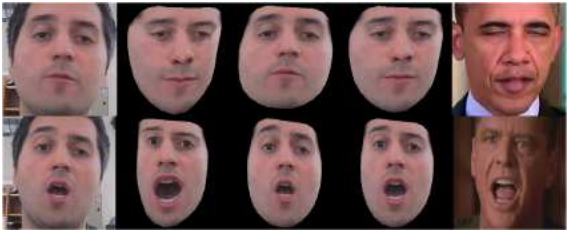}
  \caption{Comparison of warping approaches. Left: Selected user
    frame. Right: Target pose. Middle left to right: non-rigid
    warping (Eq.~\eqref{Eq:non-rigid-term}), affine warping
    (Eq.~\eqref{Eq:affine-term}), and our approach
    (Eq.~\eqref{Eq:total_energy}). 
    % Note that non-rigid warping
    % distorts eyes and mouth, while affine warping fails to find a
    % correct view deformation.
  }
  \label{Fig:warping_strategies}
\end{figure}

Note that a shape and appearance transfer as described here are
generally not possible with conventional warping approaches, such as
global non-rigid warping and global affine warping, as shown in
Fig.~\ref{Fig:warping_strategies}. The former creates unrealistic
distortions in texture since it fits the source shape exactly to the
target shape, while the latter may fail under strong perspective views
and create odd deformations whenever the source and target shape do
not agree.

\subsection{Compositing}

\begin{figure}
  \centering
    \includegraphics[height=1.6cm]{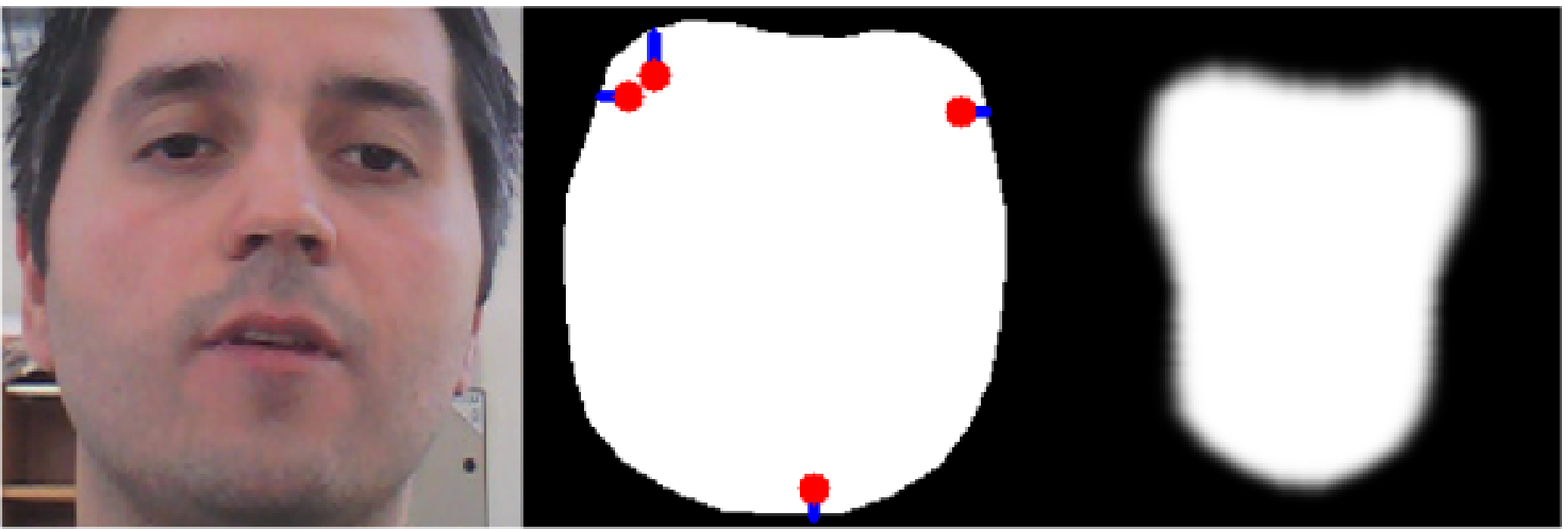} \\
    \includegraphics[width=.9\linewidth]{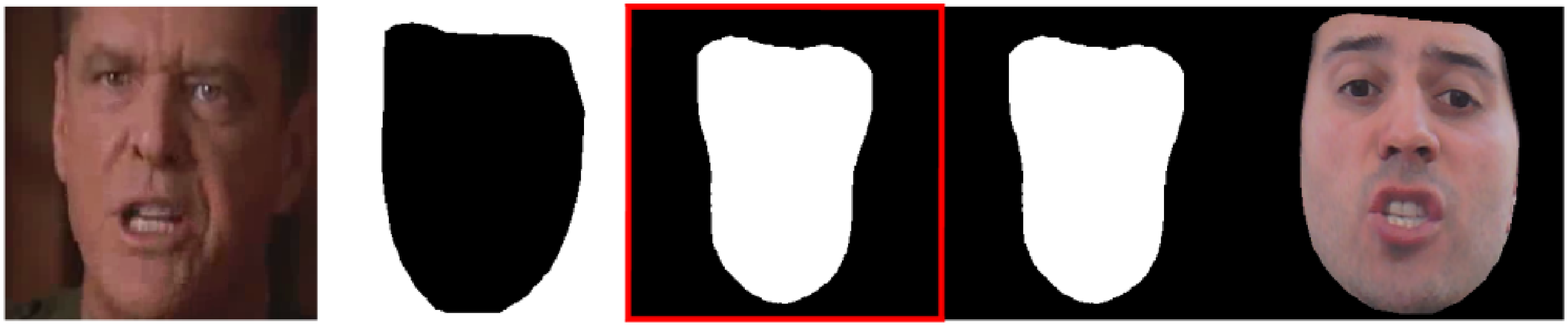} \\
    \caption{Seam generation. Top: User at rest, source mask with
      landmarks closest to the boundary in red, and eroded mask.  Bottom
      left: Target frame and mask. Bottom Right: Transferred source
      frame and mask. Bottom middle: Final blending seam.
      % obtained by
      % intersecting source and target mask.
    }
  \label{Fig:mask_generation}
\end{figure}

% Having transferred the source face to the target sequence,
We produce a convincing composite, where the main facial source
features, represented by the eyes, nose, mouth, and chin are
seamlessly implanted on the target actor. The lighting of the target
sequence, and the skin appearance and hair of the target actor, should
be preserved. For this purpose we use Poisson seamless cloning
\cite{Perez03}. We create a tight binary mask for the source sequence
containing the main facial features of the user at rest, such as eyes,
mouth, nose and eyebrows. We then perform an erosion with a Gaussian
structuring element that is constrained by the landmark locations in
the facial features.  Thresholding this mask gives us a seam for
blending (see Fig.~\ref{Fig:mask_generation}, top).

To obtain a seam for each frame in the reenactment, the precomputed
source mask is transferred by inverse-warping (see
Sec.~\ref{Sec:shape_and_appearance_transfer}). We prevent the seam
from running outside the target face by intersecting it with a mask
containing the main facial features of the target actor (see
Fig.~\ref{Fig:mask_generation}, bottom).  For increased insensitivity
to the source illumination, we transform the source and target frames
into the perception-based color space of \cite{Chong08} before
performing Poisson blending \cite{Perez03}. The blended image is
converted back to RGB space, resulting in the final composite (see
Fig.~\ref{Fig:overview}). To avoid artifacts across the seam, we blend
the boundary pixels using a Gaussian with a standard deviation of 9
pixels.

%===============================================================================

\section{Results}
\label{Sec:results}

We evaluate our method on two types of data: We use videos that were
prerecorded in a studio with an SLR camera to demonstrate the
reenactment quality on existing high-quality footage. We also reenact
faces in videos taken from the Internet using a random performance of
a user captured with a webcam. This demonstrates our system's ease of
use and its applicability to online content. Our system was
implemented in C++ and tested on a 3.4 GHz
% Intel\textregistered
% Core$^{\text{TM}}$ i5
processor.
 % with 16GB RAM.

\paragraph{Existing Video.}

\begin{figure*}
  \centering
  \includegraphics[height = 0.35\linewidth]{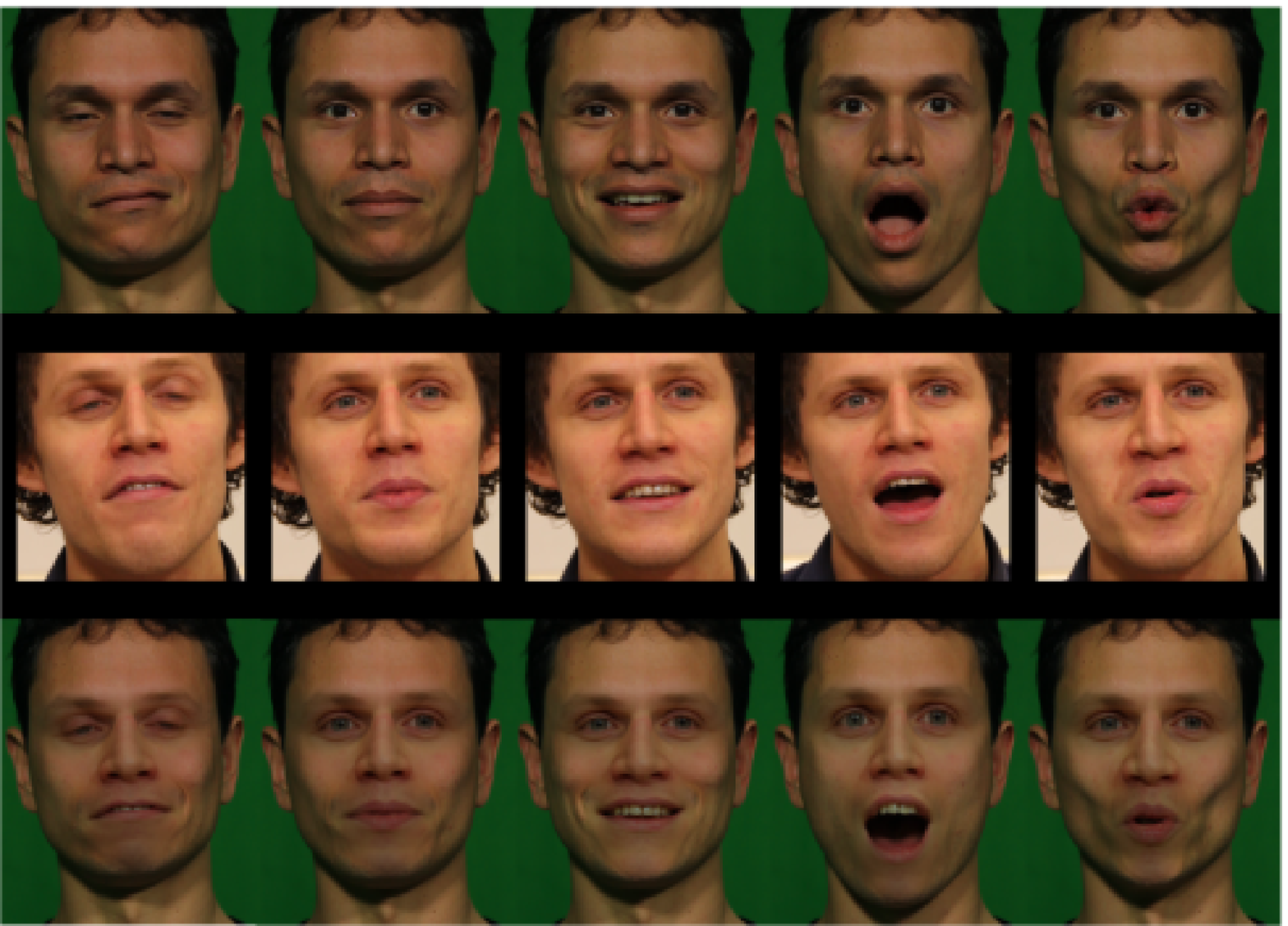} 
  \includegraphics[height = 0.35\linewidth]{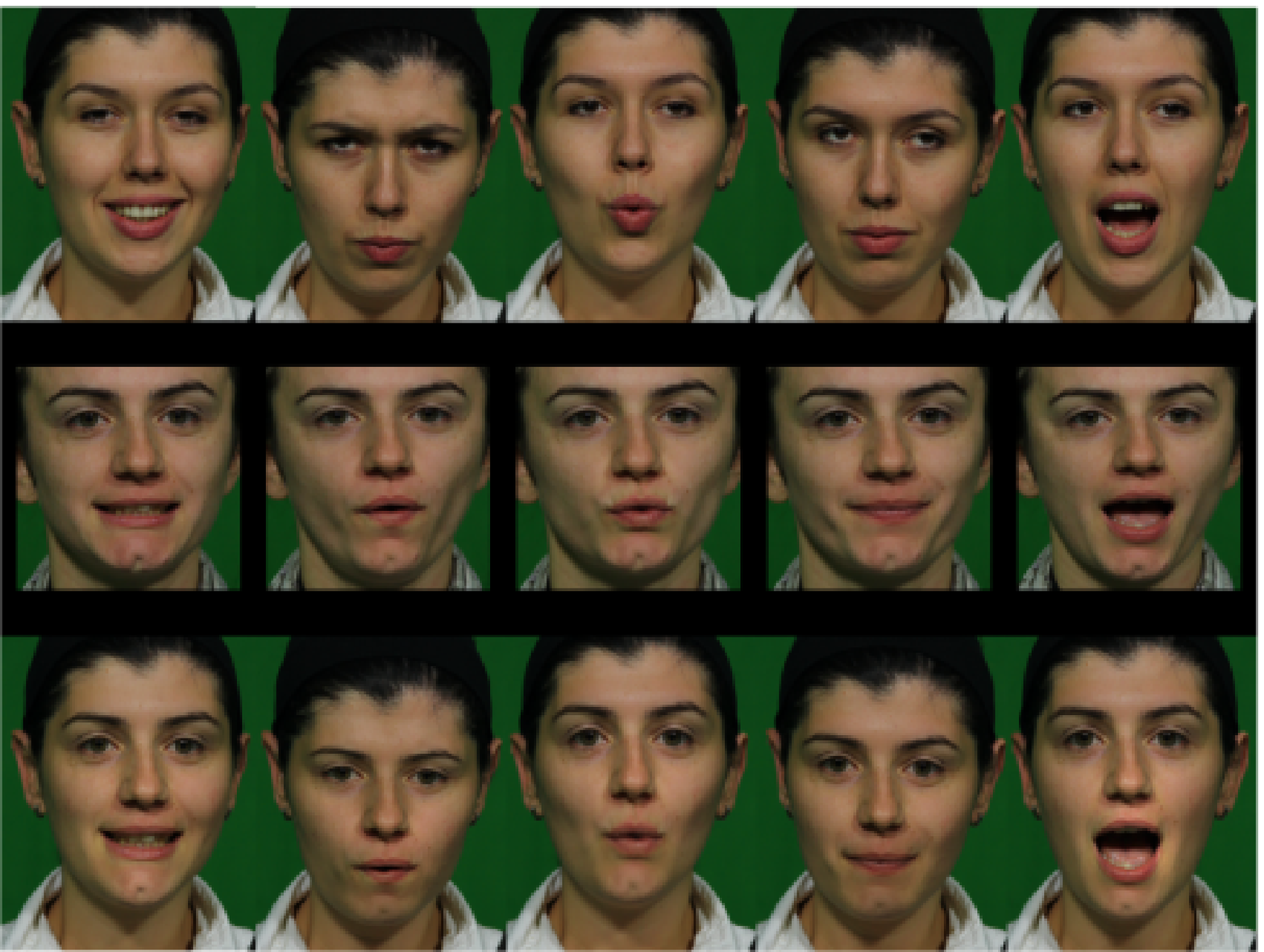}
  \caption{Existing HD video (22 s on the left, 12 s on the
    right). Top: Example frames from the target sequence. Middle:
    Corresponding selected source frames. Bottom: Final
    composite. Weights in Eq.~\eqref{Eq:total_energy}: $w_{\text{nr}}
    \! = \! 0.65$, $w_{\text{r}} \! = \! 0.35$ (left) and
    $w_{\text{nr}} \! = \!  0.55$, $w_{\text{r}} \! = \! 0.45$
    (right).}
  \label{Fig:reenactment_hd_sequences}
\end{figure*}

We recorded two male and two female users performing random facial
gestures and speech under similar ambient lighting to simulate
existing high-quality HD footage. As source sequences, we selected
% from the recordings 
a snippet of about 10 s from the first two recordings and used the
second recordings as target.
% showing one of the males and one of the females.
Fig.~\ref{Fig:reenactment_hd_sequences} shows the two reenactment
results of 22 and 12 s.
% attained by using the recordings of the two other subjects as
% target.
Note that our system is able to reproduce the target performance in a
convincing way, even when head motion, expression, timing, and speech
of user and actor differ substantially.  Computation time for the face
tracking step was about 4 s per frame, while the combined face
matching and face transfer took 4.5 min in total for both results. To
appreciate the temporal quality of these and additional results, we
refer to the supplementary video.

\paragraph{Low-Quality Internet Video.}

\begin{figure*}
  \centering
    \includegraphics[height = 0.32\linewidth]{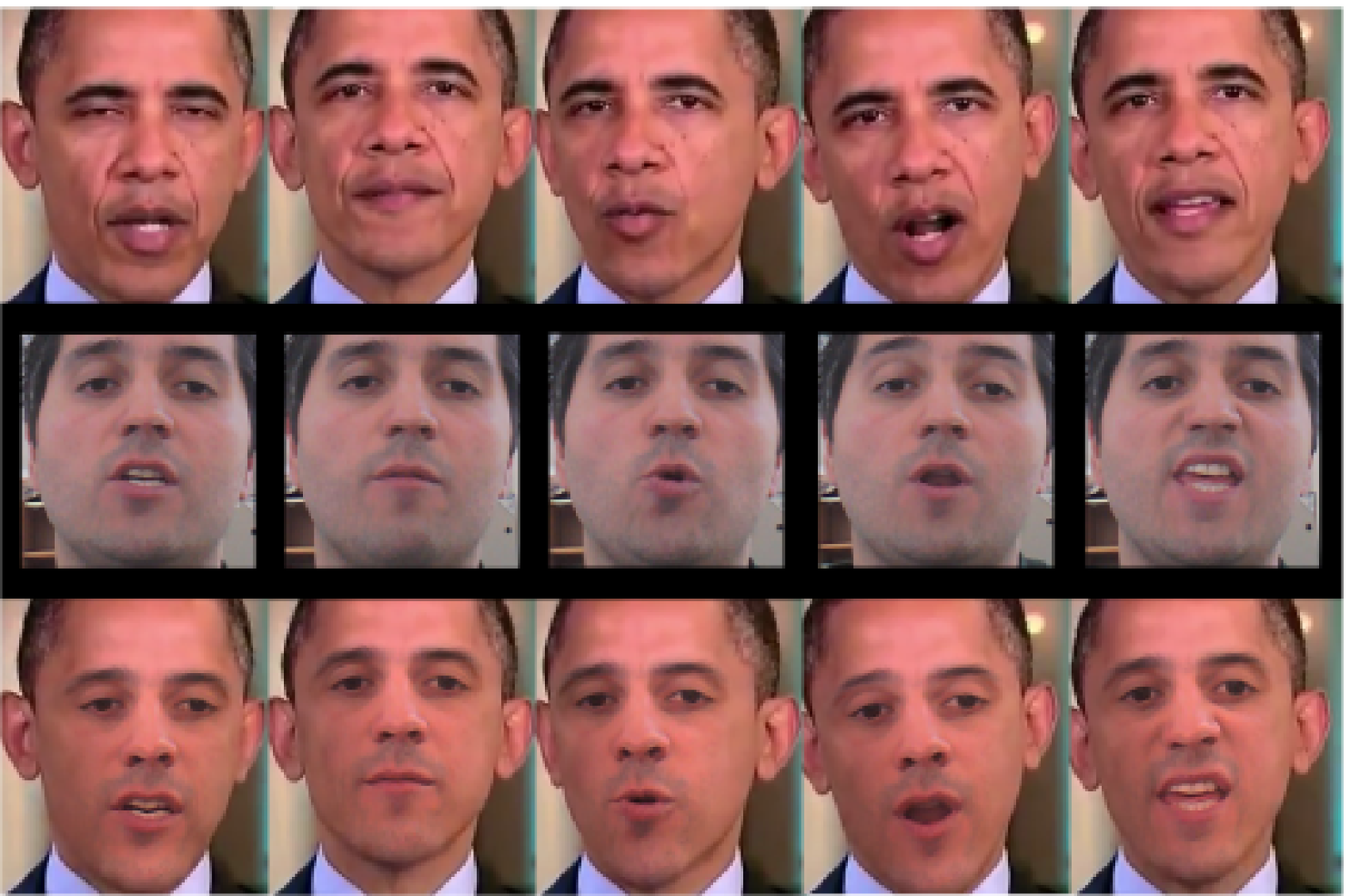} 
    \includegraphics[height = 0.321\linewidth]{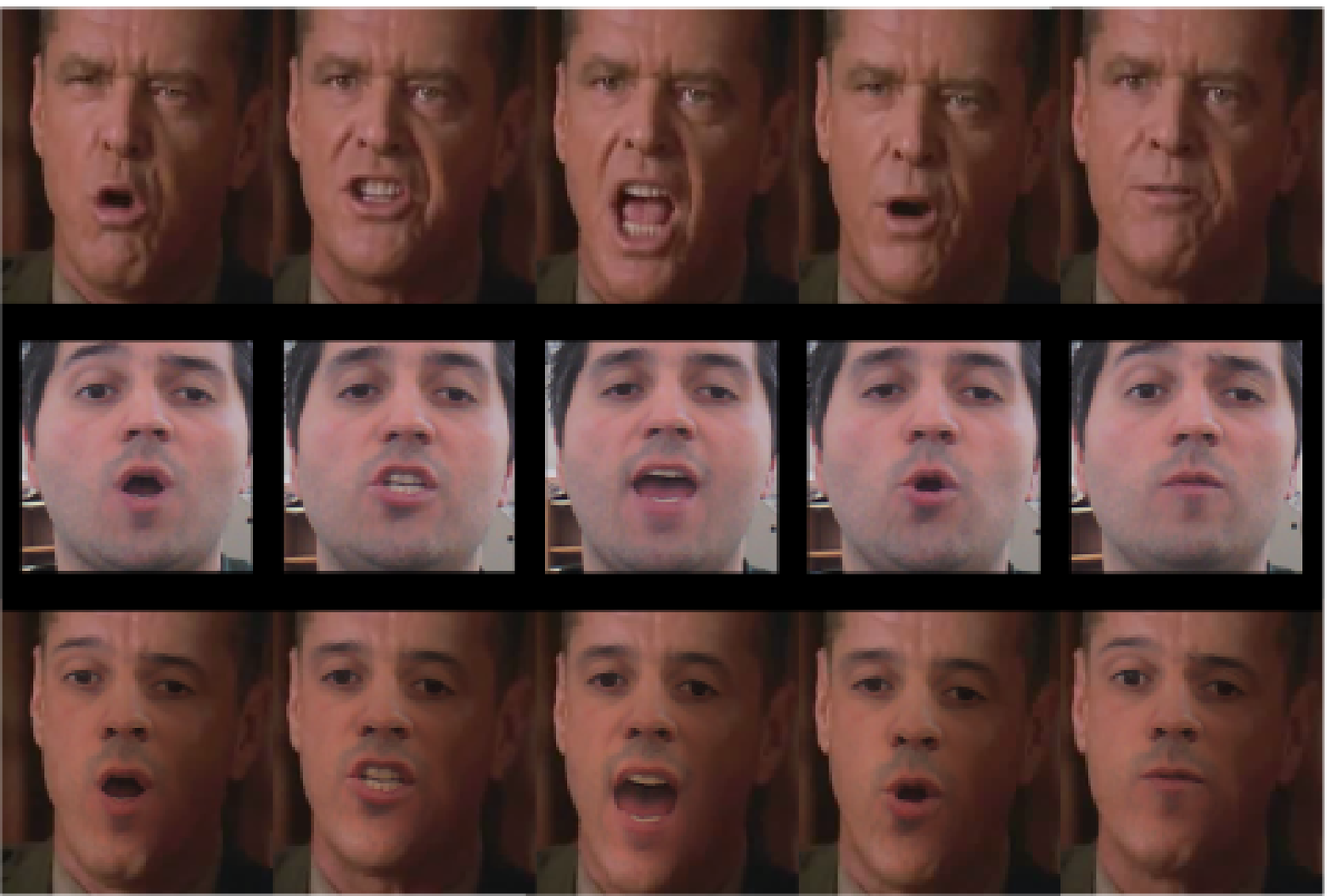}
    \caption{Low-quality Internet video (18 s from Obama's speech on
      the left (http://youtu.be/qxtydXN3f1U), 8 s excerpt from ``A Few
      Good Men'' on the right (http://youtu.be/5j2F4VcBmeo)). Top:
      Example frames from the target sequence.  Middle: Corresponding
      selected source frames. Bottom: Final composite. Weights in
      Eq.~\eqref{Eq:total_energy}: $w_{\text{nr}} \! = \! 0.65$,
      $w_{\text{r}} \! = \! 0.35$ (left) and $w_{\text{nr}} \! = \!
      0.45$, $w_{\text{r}} \! = \! 0.55$ (right).}
  \label{Fig:reenactment_youtube_videos}
\end{figure*}

Fig.~\ref{Fig:reenactment_youtube_videos} shows results for two target
videos downloaded from the Internet. The user recorded himself with a
standard webcam (20 fps, 640$\times$480) for 10 s, and the
reenactments were produced for subsequences of 18 and 8 s. Both target
videos exhibit different speech, head pose, lighting and resolution
than the recorded source sequence. Our system nevertheless produces
plausible animations, even in the presence of quite some head motion,
such as in the Obama sequence. Face matching and face transfer took
between 4 and 7 min.

\paragraph{Validation.}

A user study in the supplementary material shows with statistical
significance that our temporal clustering of
Sec.~\ref{Sec:temporal_clustering_and_frame_selection} and combined
appearance and motion distance of Eq.~\eqref{Eq:total_distance}
outperform a frame-by-frame matching with the appearance metric of
Eq.~\eqref{Eq:appearance_distance}.  Reenactment results for 5
existing and 2 web videos were rated by 32 participants w.r.t. the
original target performance in terms of mimicking fidelity, temporal
consistency and visual artifacts on a scale from 1 (not good) to 5
(good). The average scores over all results were 3.25 for our full
system, 2.92 without temporal clustering and 1.48 without motion
distance.

The supplementary material presents an analysis of the source video
length, and demonstrates that increasing the amount of source frames
improves the quality of the reenactment, both in realism and temporal
smoothness. The supplementary material also shows that reenactment and
target are almost indistinguishable when source and target are the
same sequence. For two such \emph{self-reenactments} of 10 s and 22 s,
our system produced 1 and 36 mismatches on 59 and 214 computed
clusters, respectively (a mismatch is a selected frame not contained
in the cluster). Mismatches were mostly visually very similar to the
corresponding cluster centers, such that the final reenactment is
close to a perfect frame-by-frame synthesis.  Also for the case where
source and target depict the same person under similar conditions, the
reenactment resembles the target closely.

Finally, we compared our system to the 3D approach of Dale \etal
\cite{Dale11} on data provided by the authors, depicting two different
subjects reciting the same poem. Our automatic reenactment system
produces a convincing result that is visually very close in quality to
their semi-automatic result. Since the system of Dale \etal is
designed to transfer the source face together with the complete source
performance, while our approach preserves the target performance, both
results might differ slightly in a frame-by-frame comparison (see the
supplementary video).

\paragraph{Discussion.}

Despite differences in speech, timing and lighting, our system creates
credible animations, provided that the lighting remains constant or
changes globally. Local variations can lead to wrong color propagation
across the seam and can produce flicker and less realistic
reenactments. Ghosting artifacts may also appear in the mouth region
stemming from blending and temporal inconsistencies. In the future, we
aim to drive the mouth separately, and make the compositing step more
robust to lighting changes.

Although our aim is to closely reproduce the facial expressions in the
target sequence, our reenactment results can differ from the original
performance due to the source sequence not containing a matching
expression, or the limited precision of our matching metric. Even
under perfect matching conditions, our system will preserve
person-specific nuances and subtle specialties of the source
expressions, which not only differ in detail from the target
expressions, but also between individual users of the system.

%===============================================================================

\section{Conclusion}
\label{Sec:conclusion}

We proposed an image-based reenactment system that replaces the inner
face of an actor in a video, while preserving the original facial
performance. Our method requires no user interaction, nor a complex 3D
face model. It is based on expression matching and uses temporal
clustering for matching stability and a combined appearance and motion
metric for matching coherence. A simple, yet effective, image-warping
technique allowed us to deal with moderate head motion. Experiments
showed convincing reenactment results for existing footage, obtained
by using only a short input video of a user making arbitrary facial
expressions.

%===============================================================================

\section*{Acknowledgments}
This research was supported by the ERC Starting Grant CapReal (grant agreement 
335545) and by Technicolor.

%===============================================================================

{\small
  \bibliographystyle{ieee}
  \bibliography{FaceReenactment}
}

\end{document}